\documentclass[10pt,journal,compsoc]{IEEEtran}
\usepackage{lineno,hyperref}
\usepackage{times}
\usepackage{latexsym}
\usepackage{graphicx,subfigure}
\usepackage{epstopdf}
\usepackage{amsmath, amsfonts}
\usepackage{multirow}
\usepackage{upgreek}
\usepackage{bm}
\usepackage{cite}
\usepackage{indentfirst}
\usepackage{graphicx}
\usepackage{color}
\usepackage{amssymb}
\usepackage{algorithm}
\usepackage{algorithmic}
\usepackage{arydshln}
\usepackage{pinyin}
\usepackage[marginal]{footmisc}
\usepackage{CJKutf8}
\usepackage[dvipsnames]{xcolor}
\usepackage{epsfig}

\begin{document}

\title{Chinese Idiom Paraphrasing}

\author{Jipeng~Qiang, Yang~Li, Chaowei~Zhang, Yun~Li, Yunhao~Yuan, Yi~Zhu, and Xindong~Wu, ~\IEEEmembership{~Fellow,~IEEE,}
\IEEEcompsocitemizethanks{\IEEEcompsocthanksitem J. Qiang, Yang. Li, C. Zhang, Yun Li, Y. Yuan, and Y Zhu are with School of Information Engineering, Yangzhou University, Yangzhou, Jiangsu, China.\protect\\
E-mail: \{jpqiang, cwzhang, liyun, zhuyi, yhyuan\}@yzu.edu.cn
\IEEEcompsocthanksitem X. Wu is with Key Laboratory of Knowledge Engineering with Big Data (Hefei University of Technology), Ministry of Education, Hefei, Anhui, China. \protect\\
E-mail: xwu@hfut.edu.cn 
}
}

\markboth{Journal of \LaTeX\ Class Files,~Vol.~14, No.~8, April~2019}%
{Shell \MakeLowercase{\textit{et al.}}: Bare Advanced Demo of IEEEtran.cls for IEEE Computer Society Journals}

\IEEEtitleabstractindextext{%
\begin{abstract}

Idioms, are a kind of idiomatic expression in Chinese, most of which consist of four Chinese characters. Due to the properties of non-compositionality and metaphorical meaning, Chinese Idioms are hard to be understood by children and non-native speakers. This study proposes a novel task, denoted as Chinese Idiom Paraphrasing (CIP). CIP aims to rephrase idioms-included sentences to non-idiomatic ones under the premise of preserving the original sentence's meaning. Since the sentences without idioms are easier handled by Chinese NLP systems, CIP can be used to pre-process Chinese datasets, thereby facilitating and improving the performance of Chinese NLP tasks, e.g., machine translation system, Chinese idiom cloze, and Chinese idiom embeddings. In this study, CIP task is treated as a special paraphrase generation task. To circumvent difficulties in acquiring annotations, we first establish a large-scale CIP dataset based on human and machine collaboration, which consists of 115,530 sentence pairs. We further deploy three baselines and two novel CIP approaches to deal with CIP problems. The results show that the proposed methods have better performances than the baselines based on the established CIP dataset. 

\end{abstract}

}

\maketitle

\IEEEdisplaynontitleabstractindextext

\IEEEpeerreviewmaketitle

\section{Introduction} 

\begin{CJK*}{UTF8}{gbsn}
Idiom, called "成语" (ChengYu) in Chinese, is widely used in daily communications and various literary genres. Idioms are a kind of compact Chinese expressions that consist of few words but imply relatively complex social nuances. Moreover, the Chinese idioms are often used to describe similar phenomena, events, etc, which means the idioms can not be interpreted with their literal meanings in some cases. Thus, it has always been a challenge for non-native speakers, even native speakers to recognize Chinese idioms\cite{zheng-etal-2019-chid}. For instance, the idiom "亡羊补牢"(MangYangBuLao) shown in Table 1, represents "Never be late to try", instead of its literal meaning - "To mend the fence after sheep are lost". 

In real life, if some people do not understand the meaning of idioms, we have to explain them by converting them into a set of word segments that reflect more intuitive and understandable paraphrasing. In this study, we try to manipulate computational approaches to automatically rephrase idiom-included sentences into simpler sentences (a.k.a., non-idiom-included sentences) for preserving context-based paraphrasing, and then benefit both Chinese-based natural language processing and societal applications.

Since idioms are a kind of obstacles for many NLP tasks, CIP can be used as a pre-processing phase that facilitates and improves the performance of machine translation system \cite{ho2014identifying,shao2017evaluating}, Chinese idiom cloze \cite{jiang2018chengyu,zheng-etal-2019-chid}, and Chinese idiom embeddings \cite{tan-jiang-2021-learning}. Furthermore, CIP-based applications can help specific crowds, such as children, non-native speakers, the people with cognitive disabilities, to improve their abilities of reading comprehension.

We propose a new task in this study, denoted as Chinese Idiom Paraphrasing (CIP), which aims to rephrase the idiom-included sentences into fluent, intuitive, and meaning-preserving non-idiom-included sentences. We can treat CIP task as a special paraphrase generation task. The general paraphrase generation task aims to rephrase a given sentence to another one that possesses identical semantics but various lexicons or syntax~\cite{kadotani-etal-2021-edit,lu-etal-2021-unsupervised-method}. Similarly, CIP emphasizes rephrasing the idioms of input sentences to word segments that reflect more intuitive and understandable paraphrasing. In recent decades, many researchers devoted to paraphrase generation~\cite{mckeown-1979-paraphrasing, meteer1988strategies} are struggled due to the lack of reliable supervision dataset~\cite{meng-etal-2021-conrpg}. Inspired by the challenge, we establish a large-scale training dataset in this work for CIP task.

\begin{table}
\centering
\begin{tabular}{l|c}
\hline
Input & 虽然你已经犯下了错误,但是\textcolor{red}{亡羊补牢}也为时不晚. \\
      & (Although you have made a mistake, it's not too late to mend it.)                                      \\
\hline
Output  & 虽然你已经犯下了错误,但是现在改正也为时不晚. \\
\hline
\end{tabular}
\caption{Given a Chinese idiom-included sentence, we aim to output a fluent, intuitive, and meaning-preserving non-idiom-included sentence. In the example, the idiom is marked in red.}
\label{Two_Examples}
\end{table}
\end{CJK*}

\textbf{Contributions.} This study produces two main contributions toward the development of CIP systems.

First, a large-scale benchmark is established for CIP task. The benchmark is comprised of 115,530 sentence pairs, which of 8,421 idioms. 
A recurrent challenge in crowdsourcing NLP-oriented datasets at scale-level is that human writers frequently utilize repetitive patterns to fabricate examples, leading to a lack of linguistic diversity \cite{liu2022wanli}. A new large-scale CIP dataset is created in this study by taking advantage of the collaboration between humans and machines. 

In detail, we initially divide a large-scale Chinese-English machine translation corpus into two parts (idioms-included sub-corpus, and non-idioms-included sub-corpus) by judging if a Chinese sentence contains idioms. Next, we train an English-to-Chinese machine translation (MT) system using the non-idioms-included sub-corpus. Because the training corpus for MT system does not include any idiom, MT system will not translate input English sentences to idiom-included Chinese sentences. Then, the MT system is deployed to translate English sentences of idioms-included sub-corpus to non-idioms-included sentences. A large-scale pseudo-parallel CIP dataset can be constructed by pairing the idioms-included sentences of idioms-included sub-corpus and the translated non-idioms-included sentences. Finally, we employ native speakers to validate the generated sentences and modify defective sentences if necessary.


Second, we deploy five baselines to rephrase the input idiom-included sentence. Since the constructed dataset is used as the training dataset, we treat CIP task as a paraphrase generation task. (i,ii,iii) We adopt three different sequence-to-sequence (Seq2Seq) methods as baselines: LSTM-based approach, Transformer-based approach, and mT5-based approach, where mT5 is a massively multilingual pre-trained text-to-text Transformer \cite{xue-etal-2021-mt5}. (iv) People have always used dictionaries to deal with CIP problems in the past, but the dictionaries cannot be directly incorporated into Seq2Seq methods. We propose a new method that allows networks to "attach" interpretations to idioms, which means the networks can learn the correlation capabilities between interpretations and idioms. (v) mT5 is pre-trained aiming at span masked language modeling, where consecutive spans of input tokens are replaced with a mask token and the trained model can reconstruct the masked-out tokens. The CIP problem can be dealt with by the span masked language modeling when the idioms of the sentences are masked. Specifically, a CIP sentence pair can be processed to produce a (corrupted) input sentence by replacing both the idioms of the source sentence and a corresponding target extracted from the simplified sentence. The mT5-based CIP method is fine-tuned to reconstruct the corresponding target. Experimental results show that the baselines evaluated on the constructed CIP dataset can output high-quality paraphrasing of sentences that are grammatically correct and semantically appropriate.

The constructed dataset and employed baselines that are used to accelerate this research are source-opened in Github \footnote{https://www.github.com/jpqiang/Chinese-Idiom-Paraphrasing}.

\section{Related Works}

\textbf{Paraphrase Generation.} Paraphrase generation aims to extract paraphrases of given sentences. The extracted paraphrases can preserve original meanings of the sentence, but are assembled with different words or syntactic structures \cite{mckeown-1979-paraphrasing,meteer1988strategies,zhou-bhat-2021-paraphrase}.

Most recent neural paraphrase generation methods primarily take advantage of the sequence-to-sequence framework that can achieve inspiring performance improvements compared with traditional approaches~\cite{zhou-bhat-2021-paraphrase}. A long-standing issue embraced in paraphrase generation studies is the lack of reliable supervised datasets. The issue can be avoided by constructing manually annotated paired-paraphrase datasets~\cite{kadotani-etal-2021-edit} or designing unsupervised paraphrase generation methods \cite{meng-etal-2021-conrpg}.
Differ from existing paraphrase generation research, we take our attention to Chinese idiom paraphrasing that rephrases idiom-included sentences to non-idiom-included ones. 

\textbf{Chinese Idiom Understanding.} Idiom is an interesting linguistic phenomenon in the Chinese language. Compared with other types of words, most idioms are unique in perspectives of non-compositionality and metaphorical meaning. Idiom understanding plays an important role in the research area of Chinese language understanding. Many types of research related to Chinese idiom understanding have been pushed forward that can benefit a variety of related down-streaming tasks. For example, Shao et al. \cite{shao2017evaluating} focused on evaluating the quality of idiom translation of machine translation systems. Zheng et al.\cite{zheng-etal-2019-chid} provided a benchmark to assess the abilities of multiple models on Chinese idiom-based cloze tests, and evaluated how well the models can comprehend Chinese idiom-included texts. Liu et al. \cite{liu-etal-2019-neural-based} studied how to improve essay writing skills by recommending Chinese idioms. Tan et al. \cite{tan-jiang-2021-learning} investigated the tasks on learning and quality evaluation of Chinese idiom embeddings. In this paper, we study a novel CIP task that is different from the above tasks. Since the proposed CIP method can rephrase idiom-included sentences to non-idiom-included ones, it is expected that CIP can benefit the tasks related to idiom representation and idiom translation.

\textbf{Other related tasks.} Pershina et al.\cite{pershina2015idiom} studied a new task of English idiom paraphrases aiming to determine if two idioms have alike or similar meanings. They collected idioms' definitions in dictionary and utilized word embedding modelings to represent idioms to calculate the similarity of two idioms. Qiang et al. \cite{qiang2021chinese} proposed a Chinese lexical simplification method, which focuses on replacing complex words in given sentences with simpler and meaning-equivalent alternatives. It is noteworthy that the substitutes in Chinese lexical simplification are all made up of a single word, but an idiom typically cannot be substituted by a single word to express original concepts or ideas. 

\section{Human and Machine Collaborative Dataset Construction}

\begin{figure*}[ht!]
\centering
\includegraphics[width=0.95\linewidth]{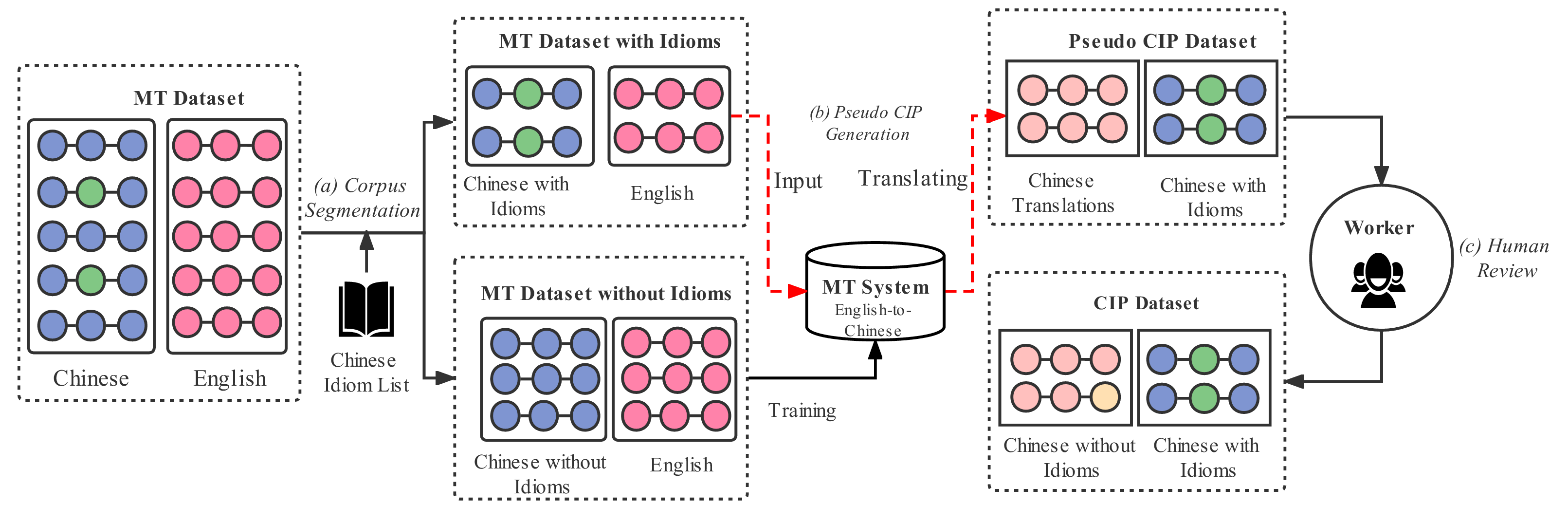}
\caption{A pipelined illustration of creating a CIP dataset based on a Chinese-English machine translation (MT) corpus. (a) The corpus is split into an idiom-included sub-corpus and a non-idiom-included sub-corpus based on a Chinese idiom list; (b) We train a MT system using the non-idiom-included sub-corpus, and create a pseudo-CIP Dataset by pairing the original Chinese idiom-included sentences and the translated non-idiom-included sentences using the trained MT system; (c) We ask human annotators to revise the translated Chinese sentence of pairs to strengthen the quality of the created CIP dataset.}
\label{pipeline}
\end{figure*}

This section describes the process of constructing a large-scale parallel dataset for Chinese idiom paraphrasing (CIP). A qualified CIP dataset needs to meet the following two requirements: (1)  The two sentences in a sentence pair have to convey the same meaning; (2) A sentence pair has to contain an idiom-included sentence and an idiom-included one. We outline a three-stage pipeline for dataset construction, which takes advantage of both the generative strength of machine translation (MT) methods and the evaluative strength of human annotators. 
Human annotators are generally reliable in correcting examples, but it is challenging while crafting diverse and creative examples at scale. Therefore, we deploy a machine translator to automatically create an initial CIP dataset, and then inquire annotators to proofread each generated sample.

\subsection{Pipeline} 

Figure 1 exhibits the details of the pipeline. Our pipeline starts with an existing English-Chinese machine translation dataset denoted as $\mathcal{D}$. Firstly, we refer a collect Chinese idiom list $\mathcal{I}$ to split the MT dataset $\mathcal{D}$ into two parts: non-idiom-included sub-dataset $\mathcal{D}1$ and idiom-included sub-dataset $\mathcal{D}2$ (Stage 1). All the data items in both $\mathcal{D}1$ and $\mathcal{D}2$ are in forms of sentence pairs. Then, we train a neural machine translation system $\mathcal{M}$ using $\mathcal{D}1$, which can translate English sentences to non-idiom-included Chinese sentences. Next, we input English sentences in $\mathcal{D}2$ to $\mathcal{M}$ to output non-idiom-included Chinese sentences. Afterwards, the Chinese sentences in $\mathcal{D}2$ and the generated sentences are mated as pairs to construct a large-scale initial parallel CIP dataset (Stage 2). Finally, the constructed dataset is reviewed and revised by annotators for quality assurances (Stage 3).

\textbf{Stage 1: Corpus Segmentation.} The English-Chinese MT dataset $\mathcal{D}$ we applied in the research are grabbed from WMT18 \cite{bojar-EtAl:2018:WMT1}, which contains 24,752,392 sentence pairs. We collect a Chinese idiom list $\mathcal{I}$ that embraces 31,114 idioms. Since the list enables determining whether the Chinese sentence in a pair contains idioms, $\mathcal{D}$ can be split as $\mathcal{D}1$ and $\mathcal{D}2$. The sub-dataset $\mathcal{D}1$ is used to train a special MT system $\mathcal{M}$ that can translate English sentences to non-idiom-included Chinese sentences. In our experiments, only 0.2\% of the translated Chinese sentences contain idioms (see Table 6). After cleansing redundant Chinese sentences, the number of sentence pairs in $\mathcal{D}2$ is 105,530. 

\textbf{Stage 2: Pseudo-CIP Dataset.} Giving a sentence pair $(c_i,e_i)$ in $\mathcal{D}2$, we input the English sentence $e_i$ into MT system $\mathcal{M}$, and output a Chinese translation $t_i$. We pair Chinese sentence $c_i$ and Chinese translation $t_i$ as a pseudo-CIP sentence pair. Thus, a CIP dataset can be built up by pairing original Chinese sentences and corresponding translated English-to-Chinese ones in $\mathcal{D}2$. The pseudo-CIP dataset $\mathcal{D}2'$ can meet the two requirements of  CIP dataset construction. On one hand, the pseudo-CIP data is from the MT dataset, which can guarantee that the paired sentences deliver the same meanings. On another hand, all original sentences include one or more idioms, and all the translated sentences do not contain idioms.

\textbf{Stage 3: Human Review.} As the final stage of the pipeline, we recruit five human annotators to review each sentence pair $(c_i,t_i)$ in the pseudo-CIP dataset $\mathcal{D}2'$. These annotators are all undergraduates, and Chinese natives. Given $(c_i,t_i)$, annotators are asked to revise and improve the quality of $t_i$. $t_i$ is required to be non-idiom-included and fully meaning-preserved.

\subsection{Corpus Statistics}

\begin{table*}[ht]
\renewcommand\arraystretch{1.5}

\centering\small

\begin{tabular}{cccccccc}
\hline
 & \multicolumn{1}{l}{} & \multicolumn{3}{c}{\textbf{In-domain}} & \multicolumn{2}{c}{\textbf{Out-of-domain}} &  \\ \hline
 & \multicolumn{1}{l}{} & \textbf{Train} & \textbf{Dev} & \textbf{Test} & \textbf{Dev} & \textbf{Test} & \textbf{Total} \\ \hline
\multicolumn{2}{c}{\textbf{sentence pairs}} & 95560 & 5000 & 5000 & 4994 & 4976 & 115530 \\ \hline
\multirow{4}{*}{\textbf{\begin{tabular}[c]{@{}c@{}}Source\\ sentence\end{tabular}}} & \textbf{token} & 3382042 & 176492 & 174753 & 225850 & 221680 & 4180817 \\ \cline{2-8} 
 & \textbf{Avg. sentence length} & 35 & 35 & 34 & 45 & 44 & 36 \\ \cline{2-8} 
 & \textbf{All Idioms} & 103168 & 5383 & 5373 & 5808 & 5801 & 125533 \\ \cline{2-8} 
 & \textbf{Unique Idioms} & 8243 & 5030 & 5038 & 5149 & 5128 & 8421 \\ \hline
\multirow{2}{*}{\textbf{\begin{tabular}[c]{@{}c@{}}Reference\\ sentence\end{tabular}}} & \textbf{tokens} & 3444405 & 179232 & 177438 & 239582 & 225452 & 4266109 \\ \cline{2-8} 
 & \textbf{Avg. sentence length} & 36 & 35 & 35 & 47 & 45 & 36 \\ \hline
\multicolumn{2}{c}{\textbf{Avg. edit disatance}} & 7.82 & 7.75 & 7.67 & 6.21 & 4.82 & 7.61 \\ \hline
\end{tabular}
\caption{\label{dataset} The statistics of CIP dataset. }

\end{table*}

The dataset $\mathcal{D}2'$ is treated as in-domain data, which contains 105,530 examples including 8,243 different idioms. $\mathcal{D}2'$ is partitioned into three parts: a training set \textbf{Train}, a development set \textbf{Dev}, and a test set \textbf{Test}. We sort the examples based on the frequencies of idioms in the corpus from high to low. We choose two samples from each of the first 5,000 idioms, and put the samples into \textbf{Dev} and \textbf{Test}, respectively.

We observe that both the \textbf{Train} and \textbf{Test} datasets come from a same distribution. However, when models are deployed in real-world applications, the inference might be performed on the data from different distributions, i.e. out-of-domain \cite{desai-durrett-2020-calibration}. Therefore, we additionally collected 9,970 sentences with idioms from modern vernacular classics, including prose and fiction, as out-of-domain data, to assess the generalization ability of CIP methods. Unlike MT corpus, these sentences have no English sentences as their references, we manually modify them to non-idiom-included sentences with the helps of Chinese native speakers. The statistical details of CIP dataset are shown in Table 2.

\begin{CJK*}{UTF8}{gbsn}

There are three significant differences between in-domain and out-of-domain data. First, the average length of sentences in in-domain data is around 35 words, while is about 45 words for out-of-domain data. Second, the average number of idioms in in-domain data is 1.07, which is lower than that of out-of-domain data (i.e., 1.17). Third, the sentence pairs in out-of-domain data need fewer modifications than that in in-domain data. In this case, a lack of linguistic diversity might be taken place due to human annotators often relying on repetitive patterns to generate sentences.

We present some examples of the idiom "深居简出" (reclusive) in the CIP dataset, shown in Table 3. The idiom "深居简出" can be rephrased with different descriptions, which displays the diversity of linguistics.

\begin{table}[]
\begin{tabular}{l|l}
\hline
\textbf{c} & 约翰并不有钱,他住在海边,\textcolor{red}{深居简出}。\\
\textbf{e} &  John is not rich, he live in a small way near the coast \\ 
\textbf{t} & 约翰并不有钱,他住在海边,很少出门。 \\ \hline

\textbf{c} & 她除了演唱外,其余时间则人\textcolor{red}{深居简出}。\\
 \textbf{e} & she seldom goes out at other times, except when she sings. \\
\textbf{t} & 她除了演唱外,其他时候很少出门。 \\ \hline
\textbf{c} & 他们崇尚非暴力、\textcolor{red}{深居简出}和远离现代社会的生活方式\\
\multirow{2}{*}{\textbf{e}} & they believe in nonviolence, simple living and little contact with the \\
& modern world.\\
\textbf{t} & 他们信仰非暴力、简单的生活,和远离现代社会的生活方式。 \\ \hline
\textbf{c} & 绝大多数的时候她都是\textcolor{red}{深居简出}，偶尔在公众场合出现。 \\
\multirow{2}{*}{\textbf{e}}  & the majority of the time she lives a secluded life, only going out \\
& occasionally. \\
\textbf{t} & 她大部分时间过着隐居的生活,偶尔在公众场合出现。\\ \hline
\textbf{c} & 我现在爱过幽静的，节俭的\textcolor{red}{深居简出}的生活。 \\
\textbf{e} & I chose now to live retired, frugal, and within ourselves. \\
\textbf{t} & 我现在爱过幽静的,节俭的退休的生活。\\ \hline
 
\end{tabular}
\caption{\label{examples} The examples contain the idiom "深居简出" in CIP dataset. \textbf{c} and \textbf{e} are a machine learning sentence pair, \textbf{t} is the CIP reference sentence of \textbf{c} generated by collaborating machine translation and human intervention.}
\end{table}
\end{CJK*}

\section{Methods}
In this section, we introduce three baselines and two proposed CIP methods to perform CIP task. The performances of the five methods are compared with each other in the aspect of idioms-included sentence rephrasing.


\subsection{Problem Formulation}

The CIP task can be defined as follows. Given a source sentence $\textbf{c}=\{c_1,...,c_j,...,c_J\}$ with one or more idioms, intending to produce a target sentence $\textbf{t}=\{t_1,...,t_i,...t_I\}$. More specifically, $\textbf{t}$ is expected to be non-idiom-included and meaning-preserved, where $c_j$ or $t_i$ refers to a Chinese character. In this study, we suppose to design a supervised method to approach this monolingual machine translation task. We adopt a Sequence-to-Sequence (Seq2Seq) framework that directly predicts the probability of the character-sequential translation from source sentences to target ones~\cite{bahdanau2014neural}, where the probability is calculated using the following equation 1:

\begin{equation}
P(\textbf{t}\mid\textbf{c})=\prod_{i=1}^{I}P(t_i\mid t_{<i},\textbf{c})
\label{sec:seq2seq}
\end{equation}
where $t_{<i}=t_1,...,t_{i-1}$.

\subsection{Seq2Seq method}
In this research, we adopt three Seq2Seq methods to handle CIP task that are LSTM-based, Transformer-based, and mT5-based models, respectively. 

(1) LSTM-based Seq2Seq methods are a basic Seq2Seq method, which uses a LSTM (Long Short-Term Memory \cite{hochreiter1997long}) to convert a sentence to a dense, fixed-length vector representation. In contrast to RNN, LSTM is helpful to deal with long sequences, but it fails to maintain the global information of the sequences.

(2) Transformer-based Seq2Seq methods\cite{vaswani2017attention} is a state-of-the-art Seq2Seq method that has been widely adopted to process various NLP tasks, such as machine translation, abstractive summarization, etc. Transformer applies a self-attention mechanism that directly models the relationships among all words of an input sequence regardless of words' positions. Unlike LSTM, Transformer handles the entire input sequence for once, rather than iterating words one by one.

(3) mT5 is a Seq2Seq method that uses the framework of Transformer. Currently, most downstream NLP tasks build their models by fine-tuning pre-trained language models~\cite{raffel2019exploring}. mT5 is a massively multilingual pre-trained language model that is implemented in a form of unified "text-to-text" to process different downstream NLP problems. In this study, we fine-tune the mT5-based approach to handle CIP task. 

\subsection{ Knowledge-based CIP method}

\begin{figure*}[ht!]
\centering
\includegraphics[width=0.85\linewidth]{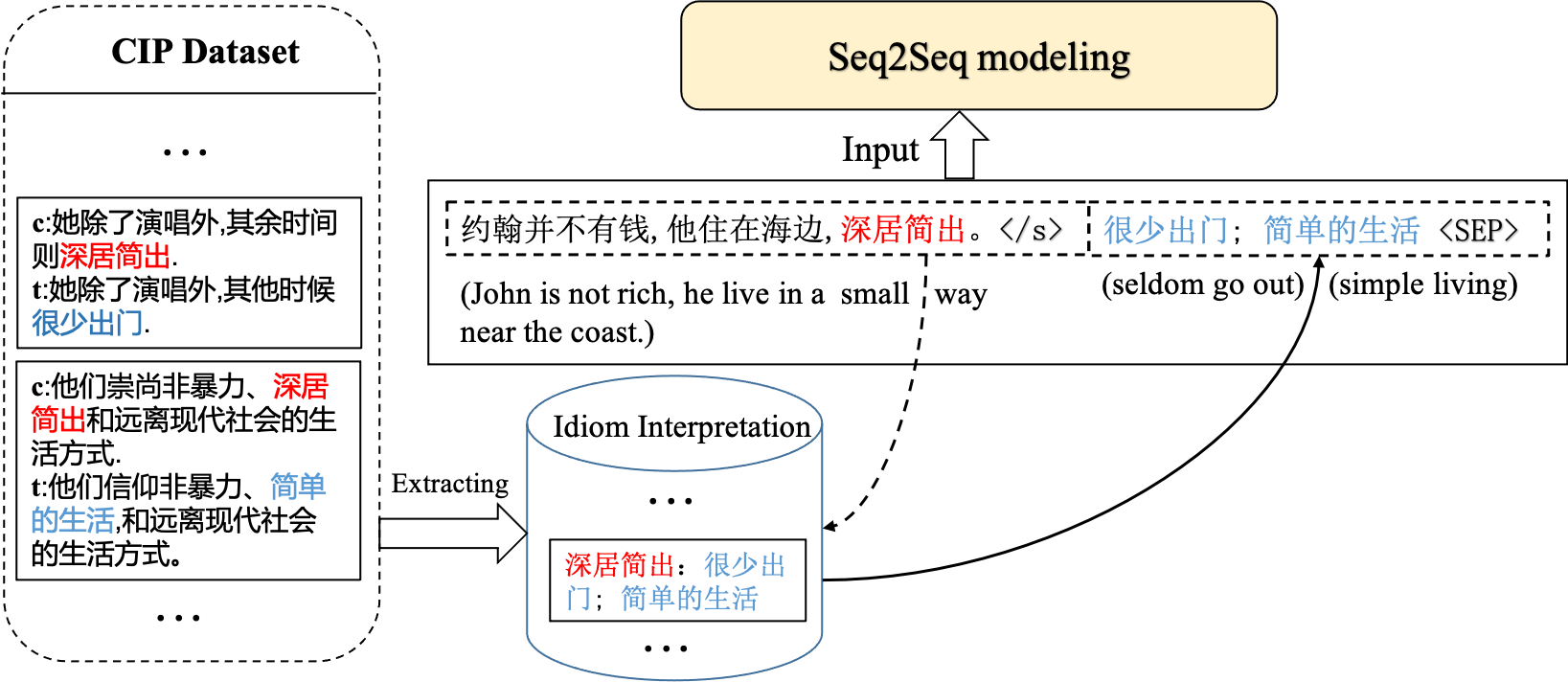}
\caption{The knowledge-based Seq2Seq method incorporates idiom interpretations into Seq2Seq modeling. The words marked in red are idioms, and the words labeled with blue are the corresponding interpretations. "$<$/s$>$" and "$<$SEP$>$" are special symbols.}
\label{knowledge}
\end{figure*}

We present an extension of Seq2Seq modeling that enables replacing idioms with their representations at the same locations in source sentences, thereby the network can better leverage idioms. The proposed method refers to a knowledge-based Seq2Seq method. Figure 2 illustrates the process of encoding a sample sentence. The model firstly extracts a dictionary of idiom interpretation from the established CIP training dataset, and then concatenates original sentences and all idioms' interpretations. Finally, the concatenations are utilized for Seq2Seq modeling.

\begin{CJK*}{UTF8}{gbsn}
(1) The interpretation of a given sentence pair (\textbf{c},\textbf{t}) is extracted by computing the edit operations. Suppose that \textbf{c} is "约翰 踢 了 我 一脚 ， 所以 我 以牙还牙 .", and \textbf{t} is "汤姆 踢 我 一脚 吧 ， 所以 我 也 踢了 他 一脚 .", where the characters of the sentence are split by spaces. We construct an edit sequences \textbf{Diff} by matching all the words in both $\textbf{c}$ and $\textbf{t}$ using three edit operations ('=','-','+'), where '=','-','+' represent 'keep', 'delete' and 'add' operations, respectively \footnote{https://github.com/paulgb/simplediff}. The output $\textbf{Diff}$ is "('-', '约翰'), ('+', '汤姆'), ('=', '踢'), ('-', '了'), ('=', '我一脚'), ('+', '吧'), ('=', '，, 所以我'), ('-', '以牙还牙'), ('+', '也踢了他一脚'), ('=', '.')". Specifically, the tuple ('-', '约翰') indicates that '约翰' appears in $\textbf{c}$ but not in $\textbf{t}$; ('+', '汤姆') denotes that '汤姆' appears in $\textbf{t}$ but not in $\textbf{c}$; and ('=', '踢') represents that '踢' is in both $\textbf{c}$ and $\textbf{t}$.

We traverse the edit sequences \textbf{Diff} using a rule $<$'-','+'$>$, where the rule means '+' follows by '-' in \textbf{Diff}. We get the following two matching sequence pairs $<$'约翰','汤姆'$>$ and $<$'以牙还牙','也踢了他一脚'$>$. A sequence pair will be ignored by the model if any idiom is included. In this example, we obtain the sequence pair $<$'以牙还牙','也踢了他一脚'$>$, where "以牙还牙" and "也踢了他一脚" represent an idiom and corresponding interpretation. Finally, we sort different interpretations of each idiom based on their frequencies, and collect the top three interpretations into the dictionary.
\end{CJK*}

(2) Given a training example $\{\textbf{c},\textbf{t}\}$, if $\{\textbf{c}$ contains multiple idioms, all interpretations of the idioms are collected from the dictionary. In case of multiple interpretations of an idiom, we first concatenate the interpretations with spaces. It is noteworthy that if a sentence has multiple idioms, we need to concatenate the interpretations of different idioms with symbol "$<$SEP$>$". Finally,  the original sentence and the collected interpretations are concatenated and processed for Seq2Seq modeling.  

We prompt the Seq2Seq modeling to learn how to make use of the concatenated information. It's noteworthy that the knowledge-based Seq2Seq method can be incorporated into all three baselines. In our experiments, we adopt mT5-based Seq2Seq modeling as the base method.

\subsection{Infill-based CIP method}

\begin{figure}[ht!]
\centering
\includegraphics[width=0.95\linewidth]{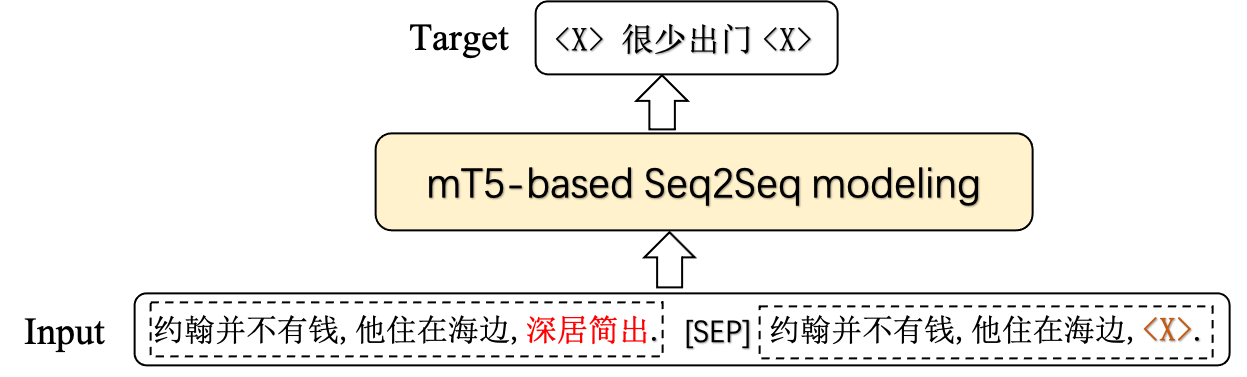}
\caption{An example of Infill-based CIP method. The input sequence fed into mT5-based Seq2Seq modeling comprised of an original sentence and a target sentence, in which the idiom of the original sentence is replaced by one special symbol "$<$X$>$". The interpretation of the idiom is treated as the reference sentence, rather than the target sentence. }
\label{infill}
\end{figure}

Given a sentence pair $\{\textbf{c},\textbf{t}\}$, we completely generate the whole target sentence \textbf{t} from the original sentence \textbf{c} using the above four CIP methods. CIP merely requires to rephrase the idioms of the sentence, which means we only expect to generate context-based interpretations of idioms, rather than the whole sentence. Therefore, we propose a novel CIP method, denoted as infill-based CIP method. 
 
As shown in Figure 3, the proposed method takes advantage of mT5 which is a pre-trained span masked language modeling (MLM) to build a Seq2Seq model. In contrast to MLM in BERT \cite{devlin2018bert}, span MLM reconstructs consecutive spans of input tokens and mask them with a mask token $<$X$>$. With the help of mT5, the proposed method enables reconstructing the idiom $I$'s interpretation of sentence \textbf{c} via replacing the idiom with the token $<$X$>$.

To embed mT5-based Seq2Seq modeling into CIP, we additionally make two modifications. (1) If the reconstructed \textbf{c} is directly fed to the span MLM, the information of idiom $I$ is likely to be ignored. We concatenate the original \textbf{c} and the reconstructed \textbf{c} as the input sequence. (2) When a sentence has multiple idioms, we recursively simplify each idiom to a corresponding representation.

\section{Experiments}
In this section, we introduce the process of conducting experiments that are used to handle CIP problem. We firstly illuminate the system configurations that include implementation details, the metrics used to evaluate performances, and baseline methods. Next, the performances of different CIP methods are assessed in Section 5.2. Then, we acquire human evaluations to survey the capabilities of deployed CIP methods on CIP processing. In addition, we calculate the proportions of idiom-included sentences that are rephrased to non-idiom-included ones. Finally, a case study is demonstrated in Section 5.5 that compares the different outputs of five methods on representing difficult Chinese idioms.

\subsection{Experiment Setups}

\textbf{Implementation Details.} In this experiment design, there are five CIP methods are deployed, including: LSTM-based Seq2Seq modeling (LSTM), Transformer-based Seq2Seq modeling (Transformer), mT5-based Seq2Seq modeling (mT5), Knowledge-based CIP method (Knowledge) and infill-based CIP method (Infill). 

We implement LSTM and Transformer methods using \emph{fairseq} \cite{ott-etal-2019-fairseq}. mT5, Knowledge, and Infill methods are  mT5-based, and are fulfilled using Huggingface transformers \cite{wolf-etal-2020-transformers}. Furthermore, the sentence tokenization is accomplished using \emph{Jieba Chinese word segmenter} and \emph{BPE tokenization}. The size of the vocabulary is set to 32K. The LSTM-based Seq2Seq method adopts Adam optimizer configured with $\beta=(0.9,0.98)$, $3e^{-4}$ learning rate, and $0.2$ dropout rate. The Transformer-based Seq2Seq method maintains the hyperparameters of the base Transformer~\cite{vaswani2017attention} (base), which contains a six-layered encoder and a six-layered decoder. The three parameters ($\beta$ of Adam optimizer, learning rate, and dropout rate) in the Transformer-based method are equivalent to these in the LSTM-based method. It's noteworthy that the learning rate is gradually increased to $3e^{-4}$ by 4k steps and correspondingly decays according to the inverse square root schedule. Knowledge and Infill methods are mT5-based Seq2Seq models. For mT5, Knowledge, and Infill, we fine-tune mT5 (base) that consists of 580M parameters \footnote{https://huggingface.co/google/mt5-base}. We train the three methods with the Adam optimizer (Kingma and Ba, 2015) and an initial learning rate of $3e^{-4}$ up to 20 epochs using early stopping on development data. The training will be stopped when the accuracy on the development set did not improve within 5 epochs. We used a beam search with 5 beams for inference. 

\textbf{Metrics.} As we mentioned above, the CIP task can be treated as a sentence paraphrasing task. Therefore, We apply four metrics to evaluate sentence paraphrasing task namely, \emph{BLEU}, \emph{BERTScore}, \emph{ROUGE-1} and \emph{ROUGE-2}. \emph{BLEU} is a widely used machine translation metric, which measures opposed references to evaluate lexical overlaps with human intervention \cite{papineni2002bleu}. \emph{BERTScore} is chosen as another metric due to its high correlation with human judgments \cite{zhang-etal-2020-enhancing}. Compared to \emph{BLEU}, \emph{BERTScore} is measured using token-wise cosine similarity between representations produced by \emph{BERT}. We measure semantic overlaps between generated sentences and reference ones using ROUGE scores~\cite{lin2004rouge}. \emph{ROUGE} is often used to evaluate text summarization.  The two metrics \emph{ROUGE-1} and \emph{ROUGE-2} refer to the overlaps of unigram and bigram between the system and reference summaries, respectively.

\textbf{Baselines.} We additionally provide two unsupervised methods that can facilitate solving the CIP problem, namely, Re-translation and BERT-CLS.

(1) Re-translation: is implemented by utilizing the back-translation techniques of machine translation methods. We first translate an idiom-included sentence to an English sentence using an efficient Chinese-English translation system, and then translate the generated English sentence using our trained English-Chinese translation system (introduced by Section 3.1) to generate a non-idiom-included Chinese sentence. The Chinese-English translation system can be easily accessed through the link\footnote{data.statmt.org/wmt17\_systems}. The trained English-Chinese translation system is a transformer-based Seq2Seq method.

(2) BERT-CLS: is an existing BERT-based Chinese lexical simplification method \cite{qiang2021chinese}. In this task, an idiom is treated as a complex word that will be replaced with a simpler word.

\subsection{Performance of CIP Method}
\label{sec:Performance_CIP}
\begin{table}[]
\newcommand{\arrarstretch}{1.5}
\centering
\begin{tabular}{llcccc}
\hline
\textbf{Method} & \textbf{Domain} & \textbf{BLEU} & \textbf{BERTS} & \textbf{ROU1} & \textbf{ROU2} \\ \hline
\multirow{2}{*}{Re-translation} & In & 29.71 & 80.47 & 57.25 & 35.62 \\
 & Out & 15.71 & 74.63 & 46.78 & 21.93 \\ \hline
\multirow{2}{*}{BERT} & In & 73.58 & 91.44 & 81.11 & 73.27 \\
 & Out & 79.65 & 94.41 & 87.78 & 83.39 \\ \hline
\multirow{2}{*}{LSTM} & In & 83.71 & 94.46 & 88.07 & 82.00 \\
 & Out & 80.14 & 93.19 & 85.86 & 79.85 \\ \hline
\multirow{2}{*}{Transformer} & In & 84.20 & 94.70 & 88.44 & 82.54 \\
 & Out & 79.58 & 93.42 & 86.23 & 80.41 \\ \hline
\multirow{2}{*}{mT5} & In & 84.57 & 94.79 & 88.48 & 82.59 \\
 & Out & 86.18 & 94.93 & 88.63 & 83.58 \\ \hline
\multirow{2}{*}{Knowledge} & In & \textbf{85.25} & 94.97 & \textbf{88.94} & \textbf{83.25} \\
 & Out & 86.29 & 94.88 & 88.40 & 83.46 \\ \hline
 \multirow{2}{*}{Infill} & In & 84.88 & \textbf{95.01} & 88.86 & 83.14 \\
 & Out & \textbf{87.19} & \textbf{95.22} & \textbf{89.05} & \textbf{84.36} \\ \hline
\end{tabular}
\caption{\label{mainResults} The performances of different methods evaluated using the metrics: BLEU, BERTS, ROUGE1, and ROUGE2, where BERTS, ROU1, and ROU2 refer to BERTScore, ROUGE-1, and ROUGE-2, respectively.}
\end{table}

Table 4 summarizes the evaluation results on our established CIP dataset. The proposed five supervised CIP methods are significantly better than two unsupervised methods (Re-translation and BERT) in perspectives of the four metrics. The results reveal that the dataset is a high-quality corpus, which can help to benefit CIP task. 

Table 4 shows that the performance of LSTM-based baseline is worsen than the other four baselines on in-domain and out-of-domain test datasets. In general, the three mT5-based CIP methods (mT5, Infill, and Knowledge) outperform the other two methods (LSTM and Transformer), which suggests that CIP methods fine-tuned on mT5 can significantly improve the CIP performance. It is observed from Table 4 that knowledge-based baseline can yield the best results on in-domain test set among three mT5-based baselines, because the incorporated knowledge is extracted from the training dataset that allows neural networks to learn how to make use of the attached knowledge. Infill baseline outperforms mT5 and Knowledge on the out-of-domain test set, despite using less input information than its counterpart Knowledge-based mT5 method. It verifies that Infill is quite effective.

\subsection{Human Evaluation}

\begin{table}[]
\begin{tabular}{llcccc}
\hline
\textbf{Method} & \textbf{Domain} & \textbf{Simp.} & \textbf{Meaning} & \textbf{Fluency} & \textbf{Avg} \\ \hline
\multirow{2}{*}{Reference} & In & 3.94 & 4.27 & 4.15 & 4.12 \\
 & Out & 3.70 & 3.70 & 3.73 & 3.71 \\ \hline
\multirow{2}{*}{Re-translation} & In & 3.03 & 2.18 & 2.69 & 2.63 \\
 & Out & 2.34 & 1.68 & 2.04 & 2.02 \\ \hline
\multirow{2}{*}{BERT} & In & 3.13 & 2.28 & 2.74 & 2.72 \\
 & Out & 2.58 & 1.95 & 2.39 & 2.31 \\ \hline
\multirow{2}{*}{LSTM} & In & 3.82 & 3.94 & 3.90 & 3.89 \\
 & Out & 3.62 & 3.54 & 3.57 & 3.58 \\ \hline
\multirow{2}{*}{Transformer} & In & 3.86 & 3.82 & 3.83 & 3.84 \\
 & Out & 3.83 & 3.74 & 3.76 & \textbf{3.78} \\ \hline
\multirow{2}{*}{mT5} & In & 3.77 & 3.85 & 3.83 & 3.81 \\
 & Out & 3.58 & 3.51 & 3.58 & 3.56 \\ \hline
\multirow{2}{*}{Knowledge} & In & 3.72 & 3.80 & 3.80 & 3.77 \\
 & Out & 3.77 & 3.71 & 3.76 & 3.75 \\ \hline
 \multirow{2}{*}{Infill} & In & 3.83 & 4.06 & 3.98 & \textbf{3.96} \\
 & Out & 3.71 & 3.79 & 3.76 & 3.76 \\ \hline
\end{tabular}
\caption{\label{humanEvaluation} The Results of human evaluation. "Simp." denotes "simplicity", "Avg" denotes "average". }
\end{table}

\begin{CJK*}{UTF8}{gbsn}
\begin{table*}
\centering
\begin{tabular}{|l|l|}
 \hline
Sentence & 如果我们无法找到内心的平和，去别处寻找无疑\textcolor{red}{缘木求鱼}。 \\
Reference & 如果我们不能在自己身上找到和平，那么到别处去寻找是毫无意义的。\\
English & if we are incapable of finding peach in ourselves, it is pointless to search elsewhere.\\
LSTM, Transformer, mT5, Infill & 如果我们无法找到内心的平和,去别处寻找无疑是徒劳的。 \\
Knowledge & 如果我们无法找到内心的平和,去别处寻找无疑是不能成功的。\\
 \hline
Sentence & 老师惩罚了考试作弊的学生\textcolor{red}{以儆效尤}.\\
Reference & 老师惩罚了考试作弊的学生起威慑作用.\\
English & the teacher made an example of the boy who copied from another student during a test.\\
LSTM, Transformer, mT5, Infill & 老师惩罚了考试作弊的学生以威慑其他人. \\
Knowledge & 老师惩罚了考试作弊的学生起威慑作用. \\
\hline
Sentence & 这次入学考试他虽\textcolor{red}{名落孙山}，但他并不气馁.\\
Reference & 这次入学考试他虽没有通过，但他并不气馁.\\
English & the teacher made an example of the boy who copied from another student during a test.\\
LSTM & 这次入学考试他虽名榜,但他并不气馁.\\
Transformer, mT5, Knowledge, Infill & 这次入学考试他虽落榜，但他并不气馁.\\
\hline
Sentence & 尽管人们对这类透支\textcolor{red}{讳莫如深}，但它却是现实存在.\\
Reference & 尽管人们对这类透支从不提及，但它却是现实存在.\\
English & the teacher made an example of the boy who copied from another student during a test.\\
Transformer & 尽管人们对这类透支提及,但它却是现实存在.\\
LSTM, mT5, Knowledge, Infill & 尽管人们对这类透支保持沉默,但它却是现实存在.\\
\hline
Sentence & 我们的错误在无情地蔓延，且缺乏公正，任何国家均\textcolor{red}{概莫能外}。\\
Reference & 我们的错误在无情地蔓延，且缺乏公正，任何国家均不能例外。\\
English & the contagion of our mistakes shows no mercy and makes no exceptions on the basis of fair play.\\
LSTM & 我们的错误在无情地蔓延,且缺乏公正,任何国家均都不例外。\\
Transformer, Knowledge & 我们的错误在无情地蔓延,且缺乏公正,任何国家都不例外。\\
mT5 & 我们的错误在无情地蔓延,且缺乏公正,任何国家都不能幸免。\\
Infill & 我们的错误在无情地蔓延,且缺乏公正,任何国家均不例外。\\
\hline
\end{tabular}
\caption{Output examples from different methods. The idioms of sentences are marked in red. }
\label{Example5}
\end{table*}

For further evaluating the CIP methods, we adopt human evaluation to analyze the deployed CIP methods. We manually select 32 infrequent and easily misused idioms. We choose 30 sentences from in-domain test set and 32 sentences from out-of-domain test set. We inquire four native speakers to rate each generated sentence using three features: simplicity, meaning, and fluency. The five-point \emph{Likert scale} is adopted to rate these features, and the average scores of the features are calculated correspondingly. (1) Simplicity is responsible for evaluating whether re-paraphrased idioms of generated sentences are easily understandable, which means idioms in original sentences should be rewritten with simpler and more common words. (2) Meaning assesses if generated sentences preserve the meaning of original sentences. (3) Fluency is used to judge if a generated sentence is fluent, and does not contain grammatical errors.

The results of human evaluation are shown in Table 5. We calculate the scores of annotated sentences \textbf{t}, denoted as Reference. We first analyze the results of these CIP methods on in-domain test set. The infill-based mT5 method outperforms other methods, which means Infill is an effective method on CIP task. The human evaluation performance of Knowledge-based CIP method is worse than that of LSTM, Transformer, and mT5. Since the knowledge-based baseline has the best results among all the baselines while adopting automatic metrics（see Table 4), it is inconsistent with the results yielded using human evaluation. We also notice that the three idioms are wrongly rephrased due to the "attached" information. If the "attached" information is fitful to this example, it may bring noise into the network.

Then, we analyze the results on an out-of-domain test set. The scores of the four CIP methods (Transformer, Infill, and Knowledge) are slightly larger than the scores of Reference. Although it is abnormal, the sentences generated by the CIP methods are certainly high-quality. The out-of-domain test set is individually collected by native Chinese crowd-workers without human and machine collaboration. The crowd-workers often take limited writing strategies to speed up the establishment of a dataset, which is harmful to the diversity of the dataset \cite{geva-etal-2019-modeling,liu2022wanli}. The quality of out-of-domain test set can be further improved. 

\subsection{Proportion of Idiom Paraphrasing}
\begin{table}[]
\centering
\begin{tabular}{lcc}
\hline
\textbf{Method} & \textbf{In} & \textbf{Out} \\ \hline
Re-translation & 99.72\% &  99.78\%\\ \hline
BERT & 81.48\% & 74.74\% \\ \hline
LSTM &  92.64\% & 85.79\%\\ \hline
Transformer & 94.22\% & 88.32\% \\ \hline
mT5 & 89.34\% & 73.45\% \\ \hline
Knowledge & 91.82\% & 85.76\% \\ \hline
Infill &  91.04\% & 84.08\% \\ \hline
\end{tabular}
\caption{\label{proportion} The proportion between the times of idiom paraphrasing and the number of all idioms.}
\end{table}

CIP aims to rephrase an input idiom-included sentence to a meaning-preserved and non-idiom-included sentence. In this subsection, we count the number of idiom-included sentences that are rephrased to non-idiom-included sentences. The results are shown in Table 7. The result shows that Re-translation achieves the best results, which can rephrase almost all idioms to non-idiom-included representations. That means, our idea on CIP dataset construction using machine learning method is feasible. Theoretically, if the trained English-Chinese machine translation method (stage 2 in the pipeline) can output high-quality results, we do not need to ask annotators to optionally revise Chinese translations. We observe that the proportions of these CIP methods (LSTM, Transformer, mT5, Knowledge and Infill) are nearly 90\%, which means they have great potential on dealing with idiom 
paraphrasing. Moreover, a tiny part of idioms cannot be rephrased, because some idioms are simple, thereby are retained in the training set. 

\subsection{Case Study}
\label{sec:case_study}

We first choose five complicated Chinese idioms that are "缘木求鱼"(climb trees to catch fish),"以儆效尤"(warn others against following a bad example),"名落孙山"(fail in official examinations),"讳莫如深"(carefully conceal mentioning), and "概莫能外"(admit of no exception whatsoever). Then, we select a sample from our CIP test set for each idiom, and analyze the output sentences generated by the five CIP baselines.
\end{CJK*}

Table 6 shows the sentences rephrased using the five CIP methods. In general, these methods trained on the established CIP dataset are capable to generate meaning-preserved and non-idiom-included sentences. In these examples, all the generated sentences are simpler than the original sentences, even if these idioms are challenging to native speakers.

\section{Conclusion}
In this paper, we propose a novel Chinese idiom paraphrasing (CIP) task, which aims to rephrase idiom-included sentences to non-idiom-included ones. The CIP task can be treated as a special paraphrase generation task and is feasible to be dealt with by adopting sequence-to-sequence(Seq2Seq) modeling. We construct a large-scale training dataset for CIP by taking the collaborations between humans and machines. Specifically, we first design a framework to construct a pseudo-CIP dataset and then ask workers to revise and evaluate the dataset. In this study, We deploy three Seq2Seq methods and propose two novel CIP methods (Infill and Knowledge) for the CIP task. Experimental results reveal that all the five methods trained on our dataset can yield good results. In our subsequent research, the proposed two CIP methods will be reserved as strong baselines, and the established dataset will also be used to accelerate the study on this topic. In the future, we will keep exploring the abilities of our CIP methods in solving the problems of Chinese idiom understanding and Chinese idiom representation.

\section*{Acknowledgement}
This research is partially supported by the National Natural Science Foundation of China under grants 62076217, 62120106008, and 61906060. 

\bibliography{CLS_TASL}  

\begin{IEEEbiography}[{\includegraphics[width=1in,height=1.25in,clip,keepaspectratio]{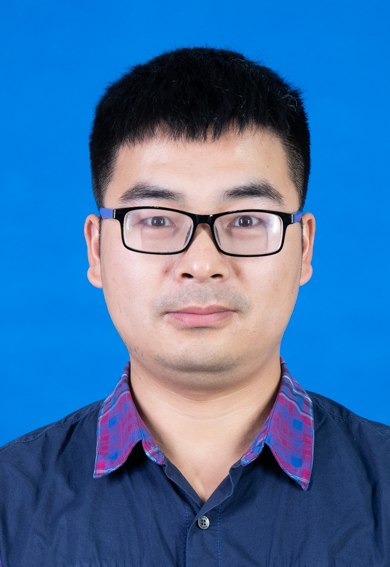}}]%
{Jipeng Qiang} is currently an associate professor in the School of Information Engineering, at Yangzhou University. He received his Ph.D. degree in computer science and technology from Hefei University of Technology in 2016. He was a Ph.D. visiting student in the Artificial Intelligence Lab at the University of Massachusetts Boston from 2014 to 2016. He has received two grants from the National Natural Science Foundation of China. He has published more than 20 papers as the first author, including AAAI, EMNLP, TKDE, TASLP, TKDD, etc. His research interests mainly include natural language processing and data mining. \end{IEEEbiography}

\begin{IEEEbiography}[{\includegraphics[width=1in,height=1.25in,clip,keepaspectratio]{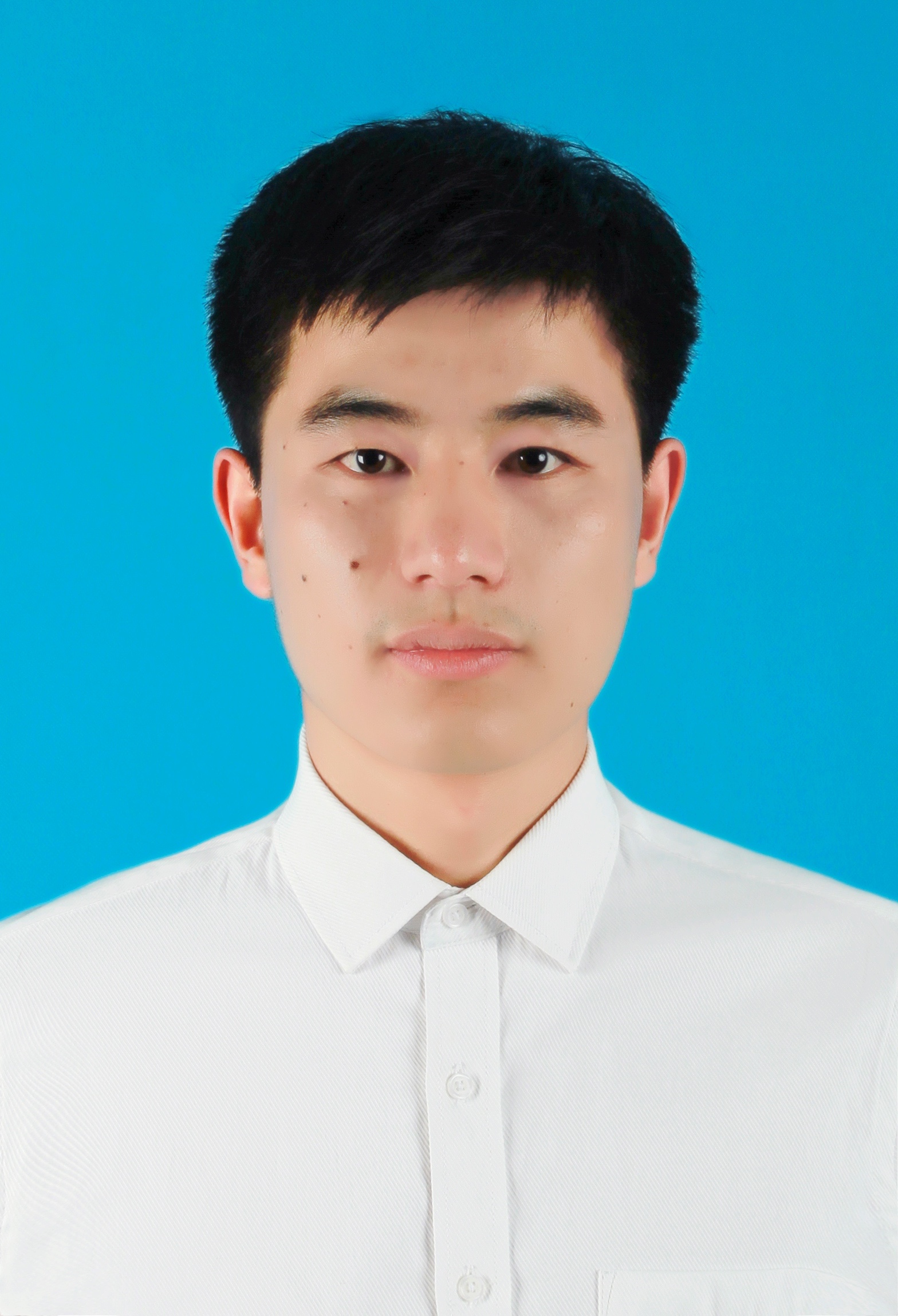}}]%
{Yang Li} is currently working toward the M.S. degree with the School of Information Engineering, Yangzhou University, Jiangsu, China. His research interest is natural language processing. \end{IEEEbiography}

\begin{IEEEbiography}[{\includegraphics[width=1in,height=1.25in,clip,keepaspectratio]{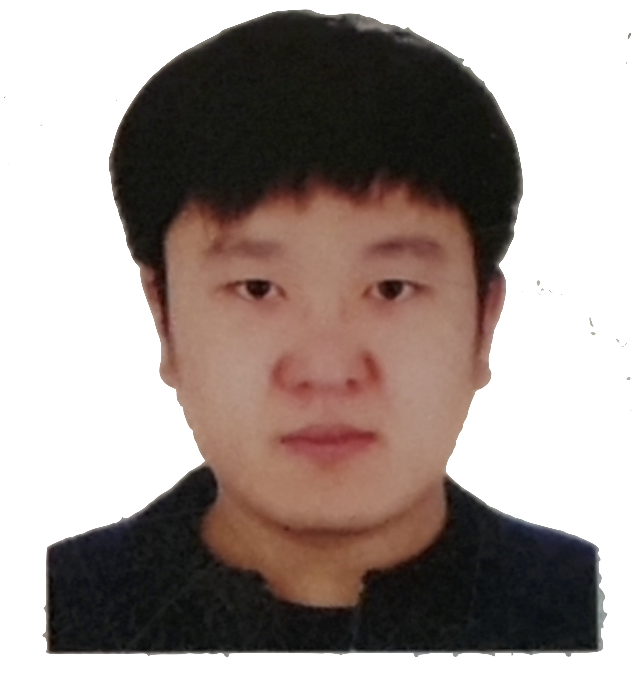}}]%
{Chaowei Zhang} received the Master \& Ph.D. degrees in computer science and software engineering from the Auburn University, USA, in 2019 and 2021, respectively. He is currently with the School of Information Engineering at Yangzhou University. His research interests include natural language processing, data mining, storage systems, and parallel computing. \end{IEEEbiography}

\begin{IEEEbiography}[{\includegraphics[width=1in,height=1.25in,clip,keepaspectratio]{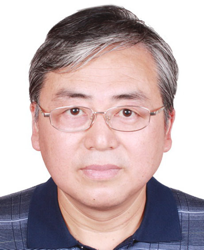}}]%
{Yun Li} is currently a professor in the School of Information Engineering, Yangzhou University, China. He received the M.S. degree in computer science and technology from Hefei University of Technology, China, in 1991, and the Ph.D. degree in control theory and control engineering from Shanghai University, China, in 2005. He has published more than 100 scientific papers. His research interests include data mining and cloud computing. 
\end{IEEEbiography}

\begin{IEEEbiography}[{\includegraphics[width=1in,height=1.25in,clip,keepaspectratio]{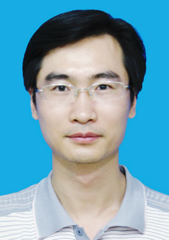}}]%
{Yunhao Yuan} is currently an associate professor in the School of Information Engineering, Yangzhou University, China. He received the M. Eng. degree in computer science and technology from Yangzhou University, China, in 2009, and the Ph.D. degree in pattern recognition and intelligence system from Nanjing University of Science and Technology, China, in 2013. His research interests include pattern recognition, data mining, and image processing. 
\end{IEEEbiography}

\begin{IEEEbiography}[{\includegraphics[width=1in,height=1.25in,clip,keepaspectratio]{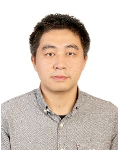}}]%
{Yi Zhu} is currently an assistant professor in the School of Information Engineering, Yangzhou University, China. He received the BS degree from Anhui University, the MS degree from University of Science and Technology of China, and the Ph.D. degree from Hefei University of Technology. His research interests are in data mining and knowledge engineering. His research interests include data mining, knowledge engineering, and recommendation systems.. \end{IEEEbiography}

\begin{IEEEbiography}[{\includegraphics[width=1in,height=1.25in,clip,keepaspectratio]{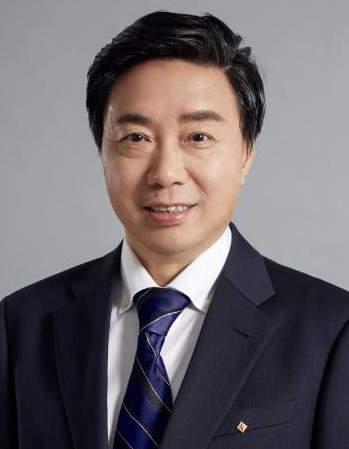}}]{Xindong Wu} is a Yangtze River Scholar in the School of Computer Science and Information Engineering at the Hefei University of Technology, China, and a fellow of IEEE and AAAS. He received his B.S. and M.S. degrees in computer science from the Hefei University of Technology, China, and his Ph.D. degree in artificial intelligence from the University of Edinburgh, Britain. His research interests include data mining, big data analytics, knowledge-based systems, and Web information exploration. He is currently the steering committee chair of the IEEE International Conference on Data Mining (ICDM), the editor-in-chief of Knowledge and Information Systems (KAIS, by Springer), and a series editor-in-chief of the Springer Book Series on Advanced Information and Knowledge Processing (AI\&KP). He was the editor-in-chief of the IEEE Transactions on Knowledge and Data Engineering (TKDE, by the IEEE Computer Society) between 2005 and 2008. He served as program committee chair/co-chair for the 2003 IEEE International Conference on Data Mining, the 13th ACM SIGKDD International Conference on Knowledge Discovery and Data Mining, and the 19th ACM Conference on Information and Knowledge Management.
\end{IEEEbiography}

\end{document}